\documentclass[conference]{IEEEtran}
\IEEEoverridecommandlockouts

\usepackage{cite}
\usepackage{amsmath,amssymb,amsfonts}
\usepackage{algorithmic}
\usepackage{graphicx}
\usepackage{textcomp}
\usepackage{xcolor}
\usepackage{multirow}
\usepackage{booktabs}
\usepackage{tabularx}
\usepackage{booktabs}
\usepackage{todonotes}
\usepackage{array} 
\usepackage{hyperref}
\def\BibTeX{{\rm B\kern-.05em{\sc i\kern-.025em b}\kern-.08em
    T\kern-.1667em\lower.7ex\hbox{E}\kern-.125emX}}

\begin{document}

\title{
Uncertainty-Aware Dynamic Knowledge Graphs \\
for Reliable Question Answering
}

\author{
Yu Takahashi$^{*}$, 
Shun Takeuchi$^{*, \dagger}$, 
Kexuan Xin$^{\dagger}$,
Guillaume Pelat$^{*}$, \\
Yoshiaki Ikai$^{*}$,
Junya Saito$^{*}$,
Jonathan Vitale$^{\dagger}$,
Shlomo Berkovsky$^{\dagger}$, 
Amin Beheshti$^{\dagger}$ \\
\small $^{*}$Fujitsu Research,
\small $^{\dagger}$Macquarie University \\
\small \{taka.yu, s.takeuchi, pelat.guillaume, ikai, saito.junya\}@fujitsu.com \\
\small \{kexuan.xin, jonathan.vitale, shlomo.berkovsky, amin.beheshti\}@mq.edu.au
}

\maketitle

\begin{abstract}
Question answering (QA) systems are increasingly deployed across domains. However, their reliability is undermined when retrieved evidence is incomplete, noisy, or uncertain. Existing knowledge graph (KG) based QA frameworks typically represent facts as static and deterministic, failing to capture the evolving nature of information and the uncertainty inherent in reasoning. 
We present a demonstration of \textbf{uncertainty-aware dynamic KGs}, a framework that combines (i) dynamic construction of evolving KGs, (ii) confidence scoring and uncertainty-aware retrieval, and (iii) an interactive interface for reliable and interpretable QA. Our system highlights how uncertainty modeling can make QA more robust and transparent by enabling users to explore dynamic graphs, inspect confidence-annotated triples, and compare baseline versus confidence-aware answers. 
The target users of this demo are clinical data scientists and clinicians, and we instantiate the framework in healthcare: constructing personalized KGs from electronic health records, visualizing uncertainty across patient visits, and evaluating its impact on a mortality prediction task. This use case demonstrates the broader promise of uncertainty-aware dynamic KGs for enhancing QA reliability in high-stakes applications. 
\end{abstract}


\section{Introduction}
Question answering (QA) systems promise to deliver interpretable reasoning across domains, but their reliability is undermined when knowledge is treated as static and deterministic. In reality, evidence is dynamic and uncertain: information changes over time, and retrieved facts can be heterogeneous, noisy, or conflicting. Without explicit mechanisms to model dynamism and uncertainty, QA pipelines risk producing brittle or overconfident answers, limiting their trustworthiness in high-stakes scenarios.

Knowledge graphs (KGs) have emerged as a way to structure heterogeneous data and support retrieval-augmented generation (RAG). By encoding entities and relations, they enhance interpretability and contextual reasoning. However, most existing graph-based QA systems still rely on static snapshots~\cite{peng2024graph, zhu2025graph} and rarely integrate explicit confidence scores on triples. Although recent work has begun exploring dynamic KG and RAG approaches~\cite{su2024dragin, thakrar2024dynagrag, rasmussen2025zep, he2025context}, practical demonstrations remain scarce. Clinical QA provides a compelling showcase: patient records are inherently longitudinal, updated across visits, and affected by uncertainty due to noise and missing information.

\begin{figure}[t]
    \centering
    \includegraphics[width=0.95\linewidth]{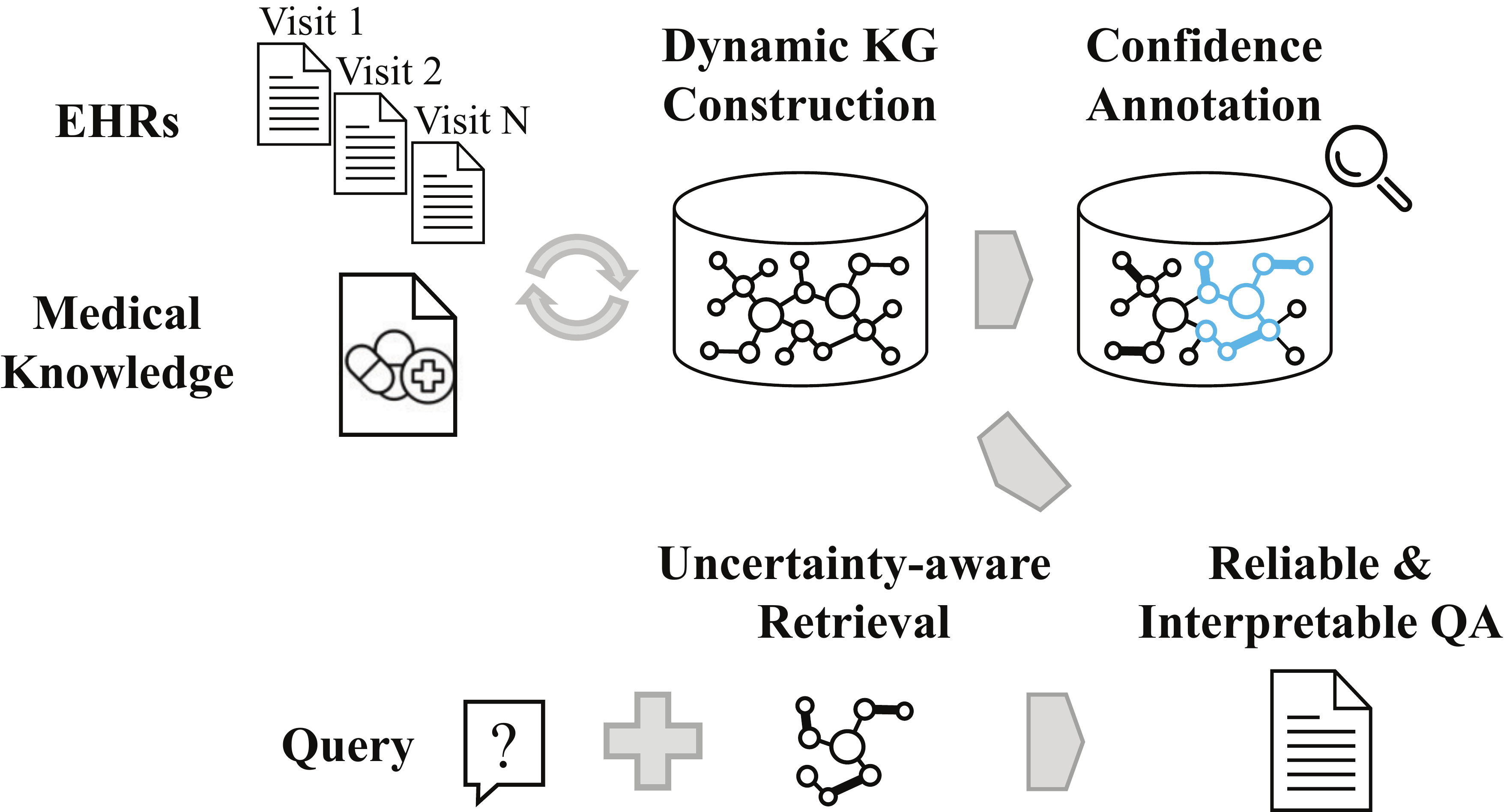}
    \caption{
    Uncertainty-aware reasoning via dynamic KGs.
    }
    \label{fig1}
\end{figure}

\begin{figure*}[t]
\centerline{\
\includegraphics[width=0.85\textwidth]{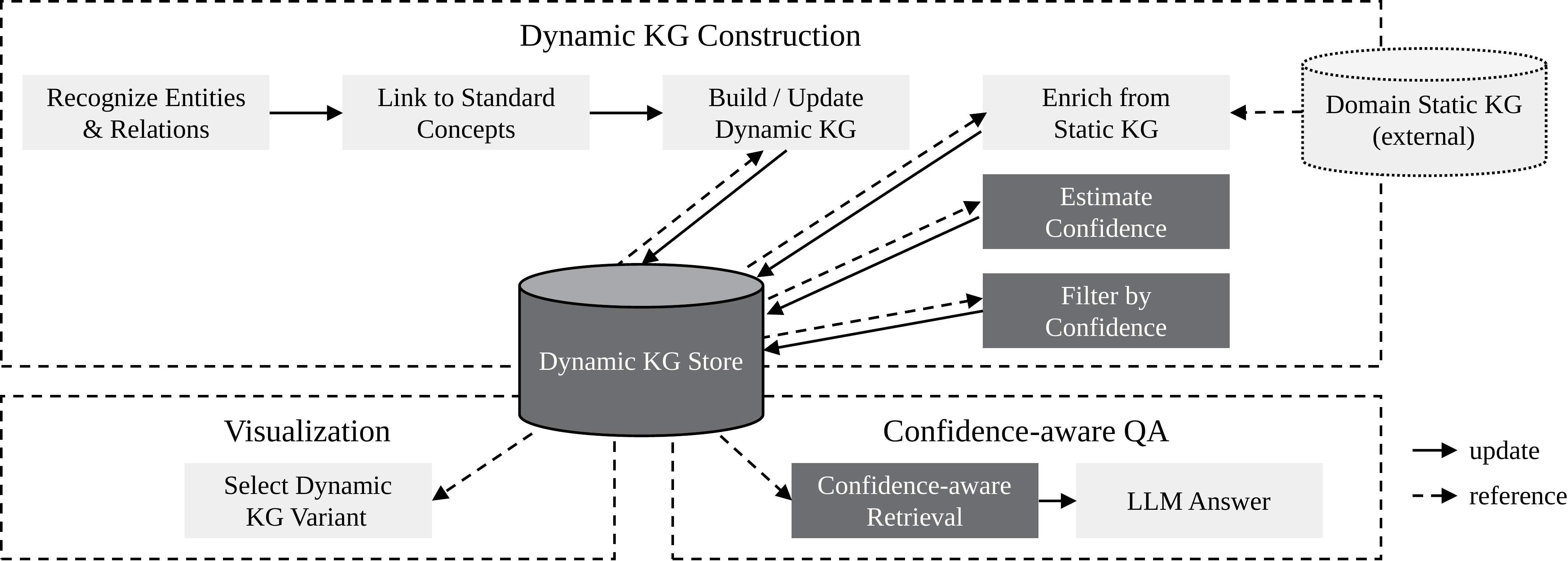}
}
\caption{
System Architecture.
The pipeline has three stages: Dynamic KG Construction, Confidence-aware QA, and Visualization.
During the Construction stage, multiple KG variants are materialized in the Dynamic KG Store 
(see Sections~\ref{sec:kg_construction} and~\ref{sec:confidence_estimation} for definitions).
Subsequent QA and Visualization stages retrieve and operate on these stored variants.
Solid arrows denote update, while dashed arrows indicate reference.
}
\label{fig:overview}
\end{figure*}

We posit that patient-centered reasoning can benefit from uncertainty-aware representation. Rather than returning answers based on raw retrieval alone, systems could incorporate confidence in clinical triples, reflecting how the available evidence supports each assertion. 
To this end, we introduce the notion of a personalized KG (PKG): a dynamic KG built for each patient individually, 
updated as the patient information evolves and annotated with confidence scores (Figure~\ref{fig1}).
We present a demo system for uncertainty-aware dynamic KGs that addresses these gaps. The system is general to QA with evolving, uncertain evidence, and we instantiate it in the healthcare domain as a validation case. The intended users of this demo are clinical data scientists and clinicians, who require interpretable and reliable QA under noisy and conflicting patient data. In practice, clinicians may employ the system as a decision-support tool during patient visits, while data scientists can use it to analyze and enhance clinical QA pipelines. Specifically, our contributions are:
\begin{itemize}
    \item \textbf{Dynamic PKG construction with evolving visit-level graphs.} Personalized graphs are incrementally updated with each patient visit.
    \item \textbf{Confidence scoring and uncertainty-aware retrieval.} 
    We compute numeric confidence scores for each triple based on their consistency with the available patient’s medical history, enabling filtering, ranking, and evidence selection that reflect uncertainty.
    \item \textbf{Healthcare showcase.} We deploy our PKG pipeline within a clinical mortality prediction task, offering an interactive interface to register electronic health records (EHRs) entries and visually explore evolving confidence-aware PKGs.
\end{itemize}

The demo provides researchers and healthcare data scientists with a testbed allowing them to explore how uncertainty shapes reasoning in clinical QA.
While KGs offer interpretability~\cite{rajabi2024knowledge}, patient-centric and confidence-aware PKGs integrate heterogeneous health data into dynamic, clinically meaningful representations, providing users with transparent and actionable insight~\cite{al2024patient}.

\section{
System Architecture and Components
}
Figure~\ref{fig:overview} shows a method-centric architecture organized into three groups: Dynamic KG Construction, Confidence-aware QA, and Visualization. The Dynamic KG Store materializes graphs during Construction, while later stages operate on the stored variants. An external Static KG provides background knowledge for enrichment. For QA, confidence-aware retrieval operates over the score-bearing KG variants (Confidence-aware or Filtered), and an LLM produces the final answer.

\subsection{Dynamic KG Construction}
\label{sec:kg_construction}
The system ingests heterogeneous text records and extracts entity and relation candidates, which are then canonicalized to a domain schema so that synonyms and surface variants map to the same concept. At each update, the system extracts new triples, which instantiate a {\bf latest} graph, which is then incrementally merged into a {\bf historical} graph that maintains longitudinal context across time. External, curated static knowledge is linked into the dynamic graph to provide background constraints (e.g., synonymy, hierarchical relations, interaction edges), yielding an {\bf enriched} variant. All graph variants produced at these stages are materialized in a Dynamic KG Store for downstream QA.

\subsection{Confidence Estimation}
\label{sec:confidence_estimation}
As a critical step within the construction pipeline, the system assigns a numeric confidence score to each triple to quantify evidential reliability. Scores reflect several signals, including source quality, repetition across time, co-occurrence structure, and temporal plausibility. They are computed using the context of the triple, which includes relevant triples both from the current and past information. This structured context is injected into the LLM prompt, and the LLM outputs a calibrated plausibility score in $[0,1]$ for the given triple. This ensures that the accumulation of evidence across records is properly accounted for.
Joint reasoning with an LLM $F$ resolves conflicts and produces a calibrated score $s$ for each triple $(h,r,t)$ (head, relation, tail) as :
\begin{equation}
s = F \bigl((h,r,t),\,\mathrm{Context}\bigr) \in[0,1].
\end{equation}
These scores yield two additional, score-bearing variants in the KG Store: a {\bf confidence-aware} graph and a {\bf filtered} subgraph obtained by thresholding low-confidence relations.

\subsection{Confidence-aware QA}
Given a natural-language query, the system retrieves relevant evidence from the KG. Two pipelines are exposed:  
\begin{itemize}
    \item \textbf{Baseline QA:} uses all triples from the KG, including current and past records, regardless of confidence.  
    \item \textbf{Confidence-aware QA:} uses either all triples or only high-confidence triples above a threshold $\tau$. In both cases, the confidence score for each triple is included:
%
\begin{equation}
T(\tau)=\{\, (h,r,t, s)\mid ((h,r,t),\,s)\in S,\ s\ge \tau \,\}.
\end{equation}
$T$ is the subset of triples, and $S$ is the set of KG triples paired with their confidence scores.
\end{itemize}

The retrieved evidence is passed to an LLM to generate an answer. By comparing baseline and confidence-aware QA, users can observe how uncertainty modeling improves reliability and reduces over-confident errors.

\begin{figure*}[t]
\centerline{\
\includegraphics[width=0.9\textwidth]
{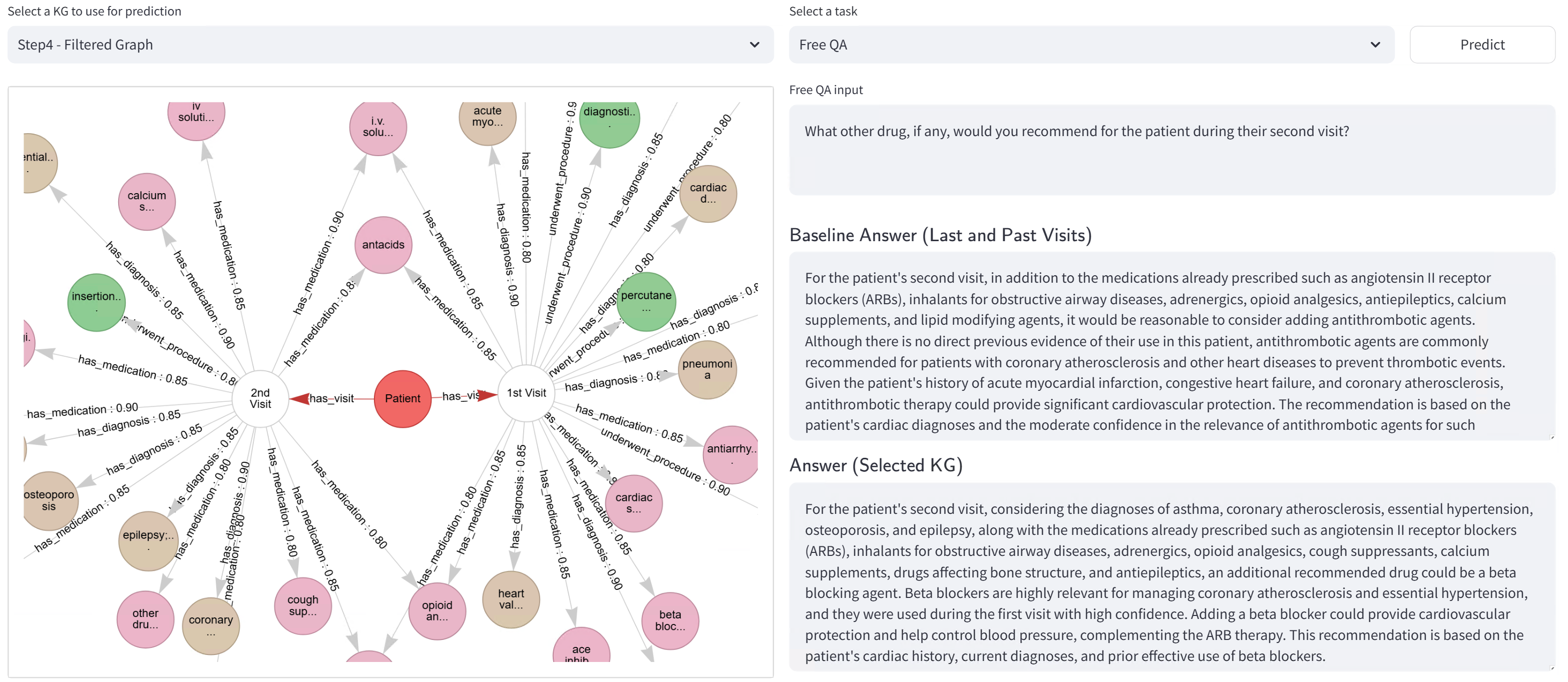}
}

\caption{
Example of clinical QA based on a filtered PKG
(patient: red, visit: white, disease: brown, procedure: green, medication: pink). The user can modify the question on the fly and compare the answers generated with different PKGs. In this case, the baseline answer suggests a generic antithrombotic drug based on the diagnoses of the patient. The Filtered KG version, which includes background knowledge and confidence, favors beta blockers citing previous high-confidence use of this drug for the patient.
}

\label{fig:CQA}
\end{figure*}

\subsection{Interactive Interface and Visualization}
The web interface exposes the pipeline end-to-end: users can enter new records, inspect how the dynamic graph evolves, and compare baseline versus confidence-aware answers. 
Multiple coordinated views enable exploration of the different KG variants (defined in Sections~\ref{sec:kg_construction} and~\ref{sec:confidence_estimation}) stored in the Dynamic KG Store.
Confidence is visualized directly on graph edges with numeric labels, and a comparative QA panel clearly highlights the differences between baseline and confidence-aware retrieval modes. The system is modular, with each stage accessible via APIs so that alternative retrievers, graph stores, or LLMs can be swapped in without altering the overall flow.
Our prototype is implemented with \textit{FastAPI} for the backend, \textit{Neo4j} for graph storage, \textit{Streamlit} for the web interface, and \textit{pyvis}~\cite{perrone2020pyvis} for interactive graph visualization.

\subsection{Healthcare Demonstration}
We instantiate the framework in clinical QA by constructing PKGs from EHRs. Textual diagnoses, procedures, and medications are canonicalized to standard terminologies (e.g., UMLS, ICD, ATC)~\cite{bodenreider2004unified} so that distinct surface forms collapse to shared concept identifiers.
Triples produced per visit form the Latest PKG, and are subsequently integrated with past patient information, external medical knowledge, and confidence annotations to produce the full set of PKG variants up to the Filtered PKG. Details are provided in Sections~\ref{sec:kg_construction} and~\ref{sec:confidence_estimation}.

Figure~\ref{fig:CQA} illustrates example visualizations and the clinical QA workflow.
The interface allows users to register new patient visits in real time, visualize patient-specific PKGs, and observe how confidence-aware retrieval alters the evidence set and final answers relative to a baseline. By providing clear, explainable visualizations of patient data structured as graphs and annotated with confidence scores, the system can help researchers identify patterns and insights that may guide improvements in model accuracy for clinical QA.
This instantiation demonstrates the method’s ability to handle longitudinal, uncertain clinical evidence while keeping the methodological core domain-agnostic.

\begin{figure}[t]
\centerline{\
\includegraphics[width=0.92\columnwidth]
{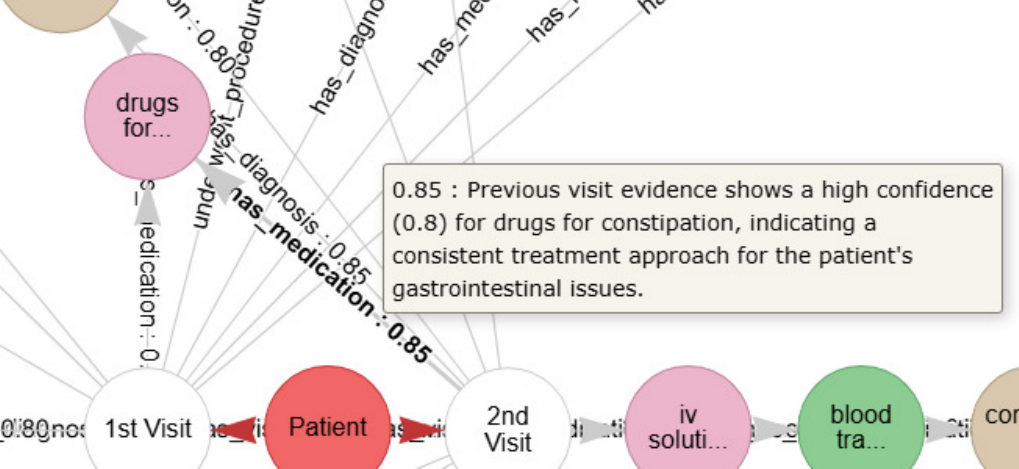}
}
\caption{
Hovering over edges allows for exploring the rationale behind the attributed confidence score, including previous evidence.
}
\label{fig:edge_conf}
\end{figure}

\section{Demonstration and Evaluation}
\subsection{Demonstration Scenario}
The demo session highlights
patient-centered QA using confidence-aware PKGs:
\begin{enumerate}
    \item \textbf{Patient selection:} The user selects a de-identified patient record.
    \item \textbf{PKG construction:} The system displays an individualized PKG with diagnoses, procedures, and medications.
    \item \textbf{Confidence display:} Triples are annotated with scores, visualized through 
    numeric badges. Hovering a triple reveals supportive/exclusive evidence and the LLM rationale, as shown in Figure~\ref{fig:edge_conf}.
    \item \textbf{Confidence filtering:} User adjusts threshold or top-k setting to retain only well-supported triples.
    \item \textbf{QA interaction:} User poses a clinical question; the system runs (i) baseline QA with all triples and (ii) confidence-aware QA with filtered/re-weighted triples. 
    \item \textbf{What-if scenarios:} Pre-defined cases show how low-confidence triples can mislead baseline QA but are suppressed in confidence-aware mode.
\end{enumerate}

\subsection{Quantitative Evaluation}
To evaluate the capability of the system on clinical QA tasks, we conducted experiments using the MIMIC-III v1.4~\cite{johnson2016mimic} EHR dataset. We focused on the mortality prediction task: given a patient's information and visit history, the goal is to predict the survival outcome at the subsequent visit. The evaluation is conducted on 1000 patients. The results presented in Table~\ref{tab:evaluation} are averaged over three runs, and we report the Area Under the Receiver Operating Characteristic Curve (AUROC) and the Area Under the Precision-Recall Curve (AUPRC), following the evaluation procedure described in~\cite{jiang2024graphcare}.

We compare three PKGs: Historical (baseline), Confidence-aware and Filtered. Predictions are made using gpt-4.1-mini.
For the Filtered PKG, the confidence threshold was varied from 0.0 to 1.0 in steps of 0.1. The value of 0.8 yielded the best performance and is reported here. Results across thresholds (omitted for brevity) consistently outperformed the baseline, confirming the robustness of the confidence-aware setting.

Table~\ref{tab:evaluation} presents the comparison in zero-shot LLM approaches for mortality prediction. 
The LLM inference with a Confidence-aware PKG outperforms the zero-shot baseline, achieving AUPRC $= 0.113$ and AUROC $= 0.522$; 6.6\% and 7.2\% improvements, respectively. Further gains are obtained with a Filtered PKG, presenting AUPRC $= 0.135$ and AUROC $= 0.575$, corresponding to 27.4\% and 18.1\% improvements compared to the baseline. These results indicate that confidence-aware retrieval contributes to higher accuracy.

\subsection{Scientific Insights from Results}
The results indicate that filtering low-confidence triples prunes noisy evidence, improving calibration and robustness while lowering semantic uncertainty, which curbs hallucinations. Triggering retrieval only when confidence is high prevents contradictions from over-retrieval and focuses grounding on truly needed facts\cite{su2024dragin}. Moreover, uncertainty-guided selection has the potential to reduce semantic uncertainty and hallucinations.

\section{Conclusion}
We addressed QA unreliability under uncertainty by pairing dynamic KGs with confidence-aware retrieval. Our method builds evolving graphs from incoming evidence, assigns numeric confidence to triples from multi-source signals, and filters the context, improving the interpretability, robustness, and trustworthiness of LLM answers. As a validation use case, we instantiated the framework in healthcare to visualize confidence-aware PKGs based on patient EHRs and to compare baseline versus confidence-aware QA, observing consistent gains and more faithful reasoning.

The approach is domain-agnostic and extends naturally to other high-stakes settings. In retail, purchase histories can be modeled as dynamic graphs to propose promotion strategies tailored to in-store purchasing behaviors. In manufacturing, incident logs can be linked to tasks, equipment, and environmental factors to recommend context-aware health and safety measures. This paradigm offers a compact path toward safer, better-calibrated LLM deployment beyond healthcare.

\begin{table}[t]
\centering
\caption{
Performance Comparison of mortality prediction.
Relative improvements over the baseline: +27.4\% (AUPRC) and +18.1\% (AUROC) with confidence-based filtering.
}
\label{tab:evaluation}
\begin{tabular}{lccc}
\toprule
Method             & PKG& AUPRC      & AUROC \\
\midrule
Baseline           & Historical & 0.106      & 0.487 \\
+ confidence score & Confidence-aware& 0.113      & 0.522 \\
+ threshold        & Filtered & \bf{0.135} & \bf{0.575} \\
\bottomrule
\end{tabular}
\end{table}

\section{Ethical Considerations}
Our work involves analysis of EHR data with sensitive health information. To ensure ethical handling, we use a private deployment of GPT-4.1-mini on Microsoft Azure, and opt out of prompt/completion storage and human review. This allows us to leverage the full capabilities of the LLM while maintaining the privacy of the data. Additionally, all used data has been de-identified in accordance with HIPAA standards.
Although on-site deployment would be ideal, it is often not feasible for many medical institutions. Thus, our setup represents a practical compromise between privacy protection and ease of use.

\bibliographystyle{IEEEtran}
\bibliography{refs}

\end{document}